\documentclass[letterpaper]{article} 
\usepackage{aaai2026}  
\usepackage{times}  
\usepackage{helvet}  
\usepackage{courier}  
\usepackage[hyphens]{url}  
\usepackage{graphicx} 
\usepackage{natbib}  
\usepackage{caption} 
\usepackage{algorithm}
\usepackage{algorithmic}
\usepackage{multirow}
\usepackage{amsmath}
\usepackage{booktabs}
\usepackage{newfloat}
\usepackage{listings}
\usepackage{tabularx}
\DeclareCaptionStyle{ruled}{labelfont=normalfont,labelsep=colon,strut=off} 
\floatstyle{ruled}
\newfloat{listing}{tb}{lst}{}
\floatname{listing}{Listing}
\usepackage{amssymb}

\title{OWL: Unsupervised 3D Object Detection by \\
Occupancy  Guided Warm-up and Large Model Priors Reasoning}
\author{
    Xusheng Guo\textsuperscript{\rm 1,\rm 2}, 
    Wanfa Zhang\textsuperscript{\rm 1,\rm 2},
    Shijia Zhao\textsuperscript{\rm 1,\rm 2}, 
    Qiming Xia\textsuperscript{\rm 1,\rm 2}, \\
    Xiaolong Xie\textsuperscript{\rm 3}, Mingming Wang\textsuperscript{\rm 4}, Hai Wu\textsuperscript{\rm 5}\corresponding, Chenglu Wen\textsuperscript{\rm 1,\rm 2}\corresponding \\
}
\affiliations{
    \textsuperscript{\rm 1}Fujian Key Laboratory of Sensing and Computing for Smart Cities, Xiamen University, China \\
    \textsuperscript{\rm 2}Key Laboratory of Multimedia Trusted Perception and Efficient Computing, \\Ministry of Education of China, Xiamen University, China  \\
    \textsuperscript{\rm 3}Tongji University, China \\
    \textsuperscript{\rm 4}Guangzhou Automobile Group Co. R\&D Center, China \\
    \textsuperscript{\rm 5}Pengcheng Laboratory, China
}

\usepackage{bibentry}

\begin{document}

\maketitle

\begin{abstract}
Unsupervised 3D object detection leverages heuristic algorithms to discover potential objects, offering a promising route to reduce annotation costs in autonomous driving. 
Existing approaches mainly generate pseudo labels and refine them through self-training iterations.  
However, these pseudo-labels are often incorrect at the beginning of training, resulting in misleading the optimization process. Moreover, effectively filtering and refining them remains a critical challenge. In this paper, we propose \textbf{OWL} for unsupervised 3D object detection by occupancy guided warm-up and large-model priors reasoning. OWL first employs an Occupancy Guided Warm-up (OGW) strategy to initialize the backbone weight with spatial perception capabilities, mitigating the interference of incorrect pseudo-labels on network convergence. 
Furthermore, OWL introduces an Instance-Cued Reasoning (ICR) module that leverages the prior knowledge of large models to assess pseudo-label quality, enabling precise filtering and refinement.  
Finally, we design a Weight-adapted Self-training (WAS) strategy to dynamically re-weight pseudo-labels, improving the performance through self-training.
Extensive experiments on Waymo Open Dataset (WOD) and KITTI demonstrate that OWL outperforms state-of-the-art unsupervised methods by over 15.0\% mAP, revealing the effectiveness of our method. 
\end{abstract}

\section{Introduction}
3D object detection aims to localize and classify objects within the environment, thereby providing essential contextual information for downstream planning and decision-making. This capability is crucial for ensuring safety in autonomous driving. With the advancement of deep learning, numerous learning-based 3D object detection approaches have emerged \cite{deng2021voxel, yin2021center, wu2023transformation, dsvt, cmd, l4dr}. However, these methods heavily rely on high-quality manual labels, and as the scope of autonomous driving expands, the reliance on ever-larger datasets makes annotation prohibitively expensive and laborious.

To reduce annotation costs while maintaining detector performance, researchers have investigated paradigms that require less annotated data \cite{guo2022hssda, hass} or utilize weaker supervisory signals \cite{ss3d, xia2023coin, hinted, zheng2025seg2box}. 
HSSDA \cite{guo2022hssda} annotates only a subset of scenes, while CoIn \cite{xia2023coin} and HINTED \cite{hinted} further minimize full-scene annotation to a single object per scene.
However, these approaches still depend on manually provided annotations as supervisory signals, and they exhibit significant performance degradation when the labeled data is either limited or of low quality.
\begin{figure}[t]
\centering
\includegraphics[scale=0.2615]{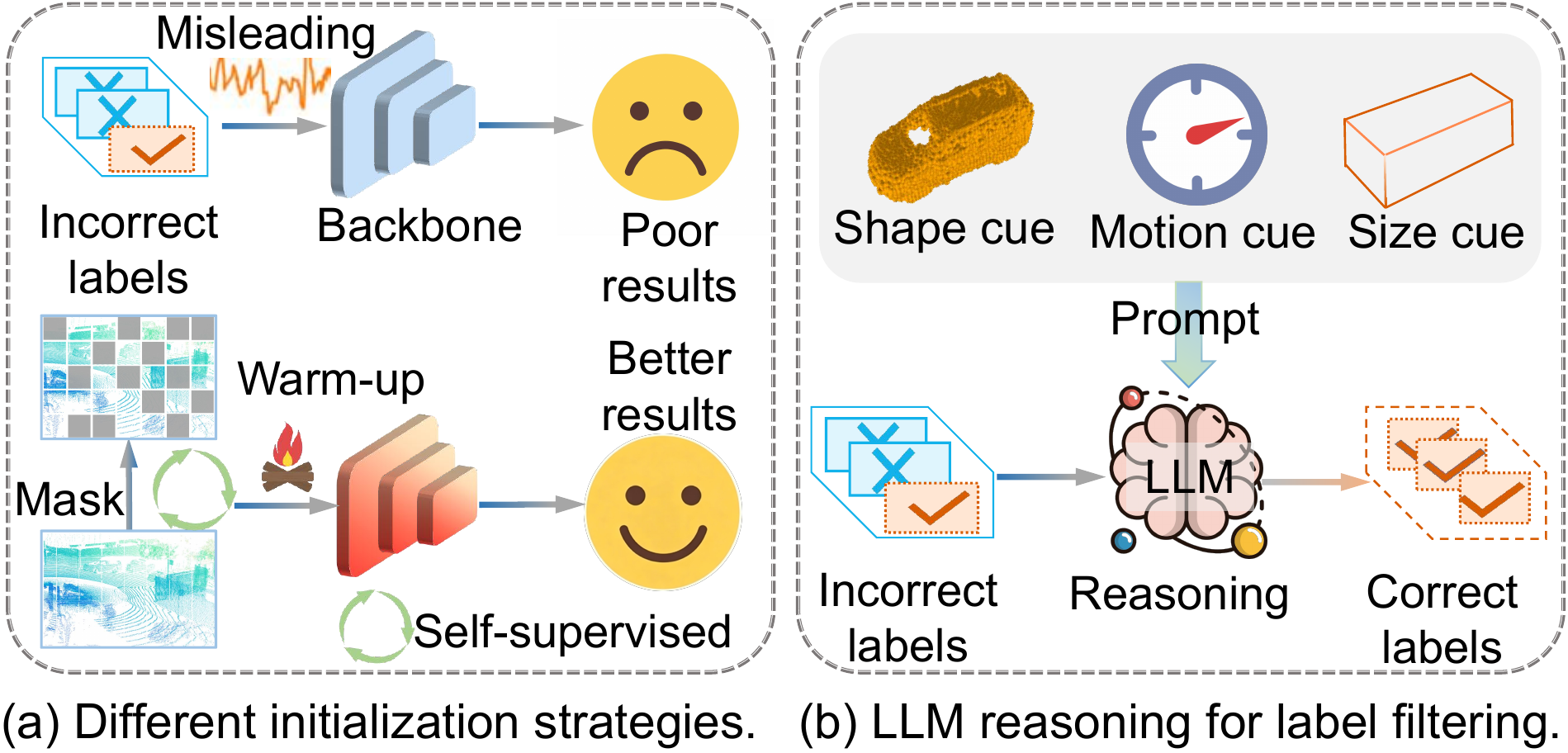} 


\caption{(a) Illustration of the effects of different backbone initialization strategies. (b) Illustration of how to refine pseudo-labels with reasoning based on instance cues.}
\label{fig1}
\end{figure}

\begin{figure}[t]
\centering
\includegraphics[scale=0.237]{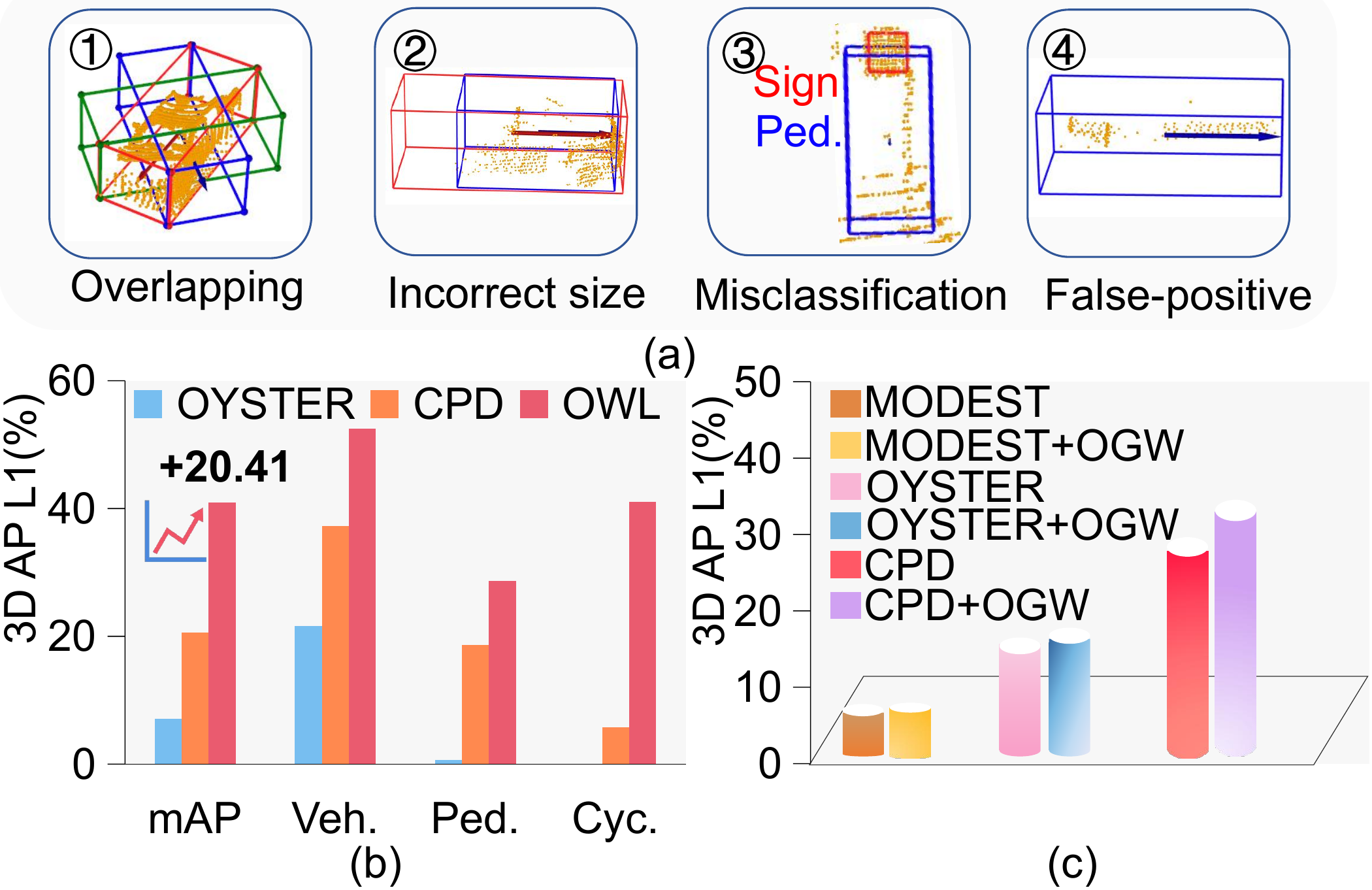} 
\caption{(a) Common errors frequently occur during the process of pseudo-label iteration. 
(b) The L1 3D AP results of OWL on the WOD test set.
(c) The impact of OGW on the performance of unsupervised methods.}
\label{fig2}
\end{figure}

Recently, some research has explored unsupervised methods that leverage commonsense knowledge priors to generate pseudo-labels instead of manual annotation \cite{you2022learning, zhang2023towards, baur2024liso, wu2024commonsense}. 
However, these unsupervised methods primarily suffer from two key challenges: (1) \emph{Unstable network initialization}. During initialization, the network is highly susceptible to inaccurate pseudo-labels, which can divert it from the correct convergence trajectory, as illustrated in Figure \ref{fig1}(a). (2) \emph{Pseudo-label filtering and refinement}. In self-training, errors in some pseudo-labels, such as misclassifications and false positives, can accumulate and amplify across iterations (see Figure \ref{fig2}(a)). 

In response to these issues, prior work has conducted relevant explorations. For the issue of initialization instability, some studies leverage self-supervised methods, such as masked encoder \cite{hess2023masked} pretraining and contrastive learning \cite{xia2023coin}, to learn more robust features.
However, these approaches are primarily designed for supervised settings and perform poorly on unsupervised tasks that lack ground-truth. 
For the second challenge, existing methods filter pseudo-labels with thresholding based on rule-specific or predictive scores \cite{wu2024commonsense, xu2023hypermodest}, which may overlook potential objects and remain error-prone.


To address these challenges,  this paper proposes an efficient framework, termed \textbf{OWL}, to improve the unsupervised 3D detection method with occupancy warm-up and large model priors reasoning. Specifically, OWL first employs an Occupancy-Guided Warm-up (\textbf{OGW})  strategy. It leverages an occupancy-prediction proxy task to capture the spatial-structural characteristics of the scene and warm up the network. This initialization strategy equips the network with a preliminary capacity for spatial feature extraction from the very beginning of training, thereby mitigating the adverse impact of erroneous pseudo-labels. Nevertheless, the strategy yields limited improvement when label errors are overly abundant (Figure \ref{fig2}(c)). Consequently, we introduce an  Instance-Cued Reasoning (\textbf{ICR}) module that harnesses large model prior reasoning. By mining instance-level cues and exploiting the extensive prior knowledge and inferential prowess of large models, the module enables accurate and efficient pseudo-label selection and refinement.  Finally, we design a Weight-adapted Self-training (\textbf{WAS}) to dynamically re-weight pseudo-labels, refining detection through iterative self-training.


Extensive experiments conducted on the KITTI and WOD datasets demonstrate that our approach surpasses both previous unsupervised and partially weakly supervised methods on the KITTI and WOD datasets, validating the efficiency of our method. The main contributions of this work are as follows:
\begin{itemize} 
\item We introduce \textbf{OWL}, which leverages an occupancy proxy task to warm up the network and harnesses the knowledge and reasoning capabilities of large models to refine pseudo-labels, substantially enhancing the performance of unsupervised 3D object detection.
\item We propose the OGW strategy for initializing the unsupervised network so that it possesses a certain perception ability at the beginning of training, mitigating the adverse influence of incorrect labels.
\item  We devise an ICR module that leverages the prior knowledge and reasoning ability of large models for pseudo-label selection and refinement, thereby markedly elevating the quality of pseudo-labels.
\end{itemize}

\section{Related Work}
\paragraph{Fully/weakly supervised 3D object detection.} 

Fully supervised 3D object detection leverages all human-annotated labels as supervisory signals to regress the bounding boxes and categories of objects within point-cloud scenes. Point-based methods \cite{shi2019pointrcnn,qi2017pointnet} extract features by directly processing the raw point cloud, whereas voxel-based approaches \cite{deng2021voxel, yin2021center, wu2022casa} partition the spatial points into voxels for subsequent computation, yielding higher efficiency at the cost of some positional fidelity. Moreover, several studies \cite{wu2023virtual, l4dr} enhance detection accuracy through multi-modal data fusion. Despite the strong performance of fully-supervised approaches, they rely on extensive annotations as supervisory signals. Since the annotation cost in full supervision is expensive, some weakly supervised works have attempted to utilize sparse annotations or semantic seed points to achieve the effect of label supervision \cite{xia2023coin,zhao2025sp3d}, reaching performance close to that of full supervision. Unlike the above, we focus more on obtaining a reliable detector without using any manual annotations.

\paragraph{Unsupervised 3D object detection.}
Unsupervised 3D detection, which does not employ real labels, faces the key challenge of generating high-quality pseudo-labels. Traditional methods that use ground point removal and point cloud clustering \cite{ester1996density} can produce preliminary pseudo-labels. In recent years, methods such as MODEST \cite{you2022learning} and OYSTER \cite{zhang2023towards} have improved pseudo-label quality by incorporating multi-pass knowledge and temporal consistency. LISO \cite{baur2024liso} and CPD \cite{wu2024commonsense} generate pseudo-labels through motion tracking and common sense prototypes, and enhance the quality of pseudo-labels via iterative self-training. Union \cite{lentsch2024union} and Motal \cite {wu2025motal} leverage multi-modal images to obtain the correct categories of objects. However, the generated pseudo-labels in these methods often contain an amount of noise and incorrect labels, which severely affect detection performance. This issue is particularly evident in iterative self-training methods. Methods like CPD and LISO iteratively filter pseudo-labels based on predicted threshold scores. However, the scores from unsupervised learning are inherently noisy, and such filtering and screening methods may miss potential objects and accumulate noise during the iterative process. Therefore, we aim to utilize more latent clues in the scene to reason the rationality of pseudo-labels and optimize the labels during the self-training process.

\begin{figure*}[!htb]
\centering
\includegraphics[scale=0.4102]{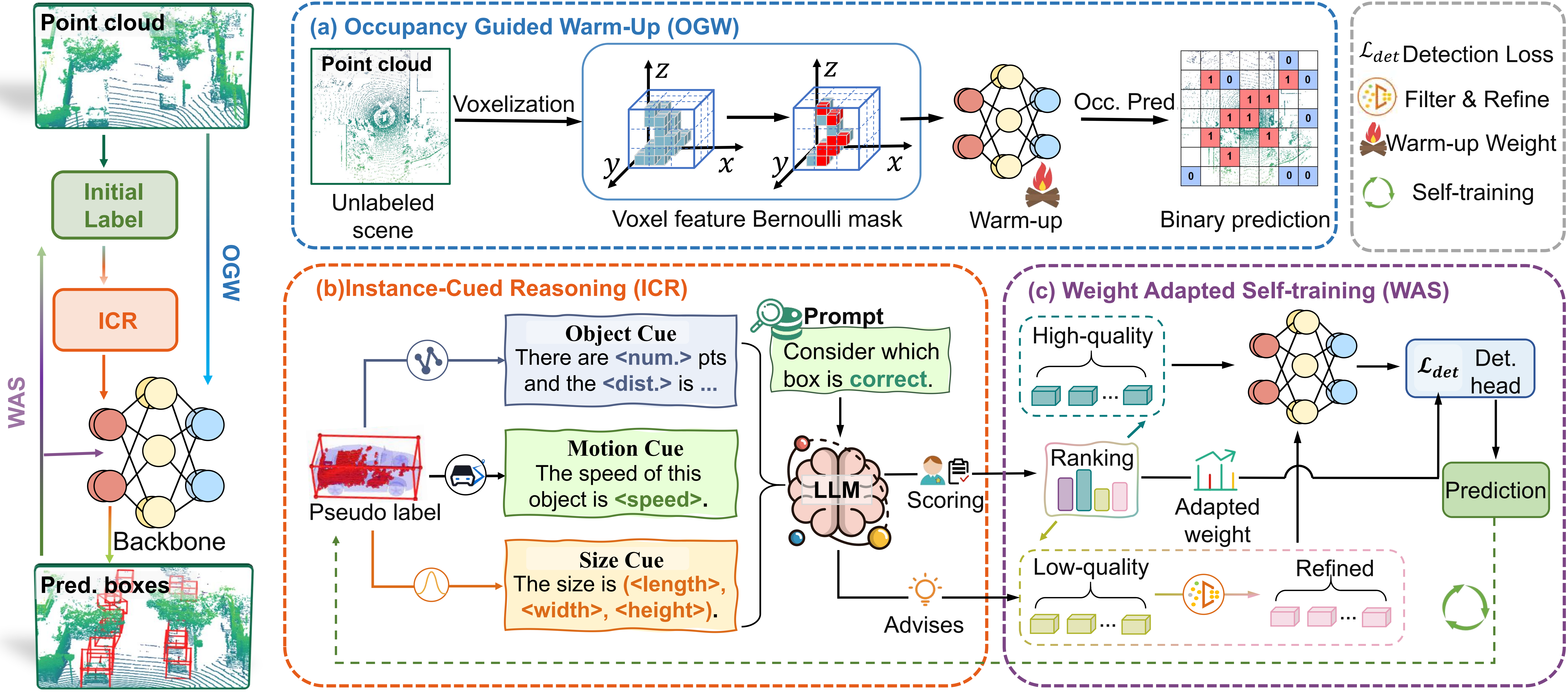} 
\label{fig3}
\caption{OWL framework. (a) Occupancy Guided Warm-Up (OGW) warms up the 3D backbone, guiding the network to learn spatial and semantic context features of the scene through the self-supervised Occupancy proxy task.
(b) Instance-Cued Reasoning (ICR) performs reasoning and judgment on various instant cues based on the knowledge of LLMs, and then filters and refines the pseudo-labels. 
(c) Weight Adapted Self-training (WAS) adaptively reweights each pseudo-label’s loss, letting the model down-weight low-confidence samples.}
\end{figure*}

\paragraph{Self-supervised Learning.}

Self-supervised learning has emerged as a pivotal paradigm by dispensing with external labels and instead exploiting inherent data structure for representation acquisition. Early endeavors \cite{gidaris2018unsupervised} learn discriminative representations via rotation prediction. Contrastive approaches such as CoIn \cite{xia2023coin} and PointContrast \cite{xie2020pointcontrast} distinguish positive and negative sample features. CPD \cite{wu2024commonsense} enhances detection capability by distilling high-quality commonsense prototypes, yet remains vulnerable to noisy pseudo-labels. Masked modeling methods including MAE \cite{he2022masked}, Voxel-MAE \cite{hess2023masked} and Occupancy-MAE \cite{min2023occupancy} refine perceptual representations through masked-token reconstruction. Prior unsupervised methods derive features exclusively from noisy pseudo-labels; to remedy this, we introduce the CRW strategy for initializing the network, mitigating the adverse influence of incorrect labels.
\section{Methodology}


\subsection{Initial Label Generation}




\paragraph{Motion Artifact Removal.} MAR \cite{wu2024commonsense,zheng2025seg2box} concatenates the continuous $2n+1$ LiDAR sweeps $\{f_{-n}, \dots, f_{n}\}$ into the current frame $f_{0}$. Foreground and background points are separated by computing the Persistence Point Score (PP-Score) \cite{you2022learning} between successive frames; only the points from the current sweep $P_{0}$ and the static points from the other sweeps are retained. After ground removal, this yields the dense point cloud $P_{0}^{*}$.

\paragraph{Dynamic Radius Clustering.} DBSCAN \cite{ester1996density} methods employed a fixed clustering radius, which can lead to inaccurate classification boundaries. Because distant point clouds are sparser and contain fewer instances, the clustering radius must be adjusted accordingly. We thus adopt a dynamic clustering radius:
\begin{equation}
    r =\alpha\cdot\bigl(1+\beta\cdot e^{-\rho}\bigr)\cdot r_0,
\end{equation}
where $\alpha$, $\beta$ denote the scaling factor and density–sensitive factor, $\rho$ denotes the density at the radius midpoint, and $r_0$ denotes the initial preset radius.
We then fit bounding boxes using the dynamic clustering radius to train the initial detector. Finally, we feed the point cloud into the detector for inference and obtain the initial pseudo-labels $\mathcal{B} $ after NMS-Selection \cite{zheng2025seg2box}.

\subsection{Occupancy Guided Warm-Up}



Previous unsupervised methods directly train on pseudo-labels, where the label noise degrades the network’s feature learning and accumulates during self-training. To learn robust feature representations, we warm up the network weights by a self-supervised strategy.

Specifically, given a dense scene $P_{0}^{*}$, we first apply voxelization, a standard preprocessing step in 3D detection, to convert the scene into a voxel grid \( V^{W \times H \times D} \) and thereby markedly enhance computational efficiency. Since the density of point clouds varies with distance, we adopt a strategy similar to Occupancy-MAE\cite{min2023occupancy}, applying a random voxel mask from near to far for each voxel. For each voxel, we calculate the distance from the voxel center to the lidar origin by
\begin{align}
(x_c,y_c,z_c)&=\frac{1}{N}\sum_{i=1}^{N}(x_i,y_i,z_i), \\
d&= \bigl\|(x_c,y_c,z_c)\bigr\|_2 .
\end{align}
Considering the disparity in the number of foreground and background points and the variation in point cloud density with distance, we employ a dynamic mask strategy as:
\begin{equation}
    p(d,w)= w \cdot (0.1 + 0.5 \cdot e^{-0.25 \cdot \lfloor d/10 \rfloor}),
\end{equation}
where \(x, y, z\) denote the point cloud coordinates, \(d\) is the distance from the voxel center to the coordinate origin, \(N\) is the number of points in the cloud, and \(p\) is the sampling ratio. Specifically, when the region's point cloud is within the bounding box of the pseudo-label \(\mathcal{B} \) (potential foreground), \(w = w_{fr} = 1\); otherwise, \(w = w_{bg} = 0.5\).
Then, we perform Bernoulli sampling based on the ratio:
\begin{equation}
\mathcal{M}\sim\text{Bernoulli}\bigl(p(d)\bigr) = \{m_j\}\in\{0,1\}^{H\times W\times D},
\end{equation}
where $\mathcal{M}$ denotes the set of voxel masks, and each $m \in \mathcal{M}$ is a binary mask for an individual voxel.

 For voxels that are not masked, they serve as the training data. The training supervision ground truth is the occupancy status of each voxel \( T \in \{0,1\}^{N_f} \), and we conduct occupancy prediction training to predict the probability of each voxel being occupied as:
\begin{equation}
    \hat{y}=f_\theta(\mathbf{V}\odot \mathcal{M})\in\{0,1\}^{H\times W\times D},
\end{equation}
where $N_f$ denotes the number of voxels not masked out, $T$ represents the voxel occupancy status, $f_{\theta}$ stands for the backbone network, and $\hat{y}$ signifies the predicted occupancy probability.

 Finally, we calculate the occupancy loss $\mathcal{L}_{occ}$. In this way, we essentially warm up the 3D backbone of the detector, enabling it to learn noise-free features and better detect incomplete and distant objects, thereby guiding the learning of noisy unsupervised labels more effectively.
\begin{equation}
    \mathcal{L}_{occ} =
-\frac{1}{|\Omega|}
\sum_{(h,w,d)\in\Omega}
\Bigl[y\log\hat{y}+(1-y)\log(1-\hat{y})\Bigr].
\end{equation}

\subsection{Instance-Cued Reasoning for label refinement}

Self-training is a crucial process for improving the accuracy of weakly supervised and unsupervised methods. However, we observe that performance plateaus after a certain number of self-training iterations. This is because errors in pseudo-labels accumulate over successive iterations, and relying solely on the OGW module is insufficient to correct these labeling errors. As shown in Figure \ref {fig2}(a), the detector is indeed capable of discovering latent objects; however, the resulting pseudo-labels often suffer from the following issues:  (1) Incorrect orientations or overlapping bounding boxes. (2) Inaccurate bounding boxes.  (3) Misdetections—including wrong categories and false positives. Previous approaches typically filter pseudo-labels based solely on the model’s predicted confidence, which risks discarding potential foreground objects. To address this limitation, we introduce the Instance-Cued Reasoning (ICR) module that leverages the prior knowledge and reasoning ability of large models for pseudo-label selection and refinement.

\begin{table*}[htbp]
\centering
\setlength{\tabcolsep}{0.35mm}
\begin{tabularx}{\textwidth}{@{}c|c|cc|cc|cc|cc@{}}
\toprule
{ } &
  { } &
  \multicolumn{2}{c|}{{ Vehicle 3D AP@0.7}} &
  \multicolumn{2}{c|}{{ Pedestrian 3D AP@0.5}} &
  \multicolumn{2}{c|}{{ Cyclist 3D AP@0.5}} &
  \multicolumn{2}{c}{{ mAP}} \\ 
\multirow{-2}{*}{{ Method}} &
  \multirow{-2}{*}{{ Reference}} &
  { L1} &
  { L2} &
  { L1} &
  { L2} &
  { L1} &
  { L2} &
  { L1} &
  { L2} \\ \midrule
{ MODEST\cite{you2022learning}} &
  { CVPR2022} &
  { 6.46} &
  { 5.80} &
  { 0.17} &
  { 0.10} &
  { 1.14} &
  { 1.10} &
  { 2.59} &
  { 2.33} \\
{ OYSTER\cite{zhang2023towards}} &
  { CVPR2023} &
  { 14.66} &
  { 14.10} &
  { 0.18} &
  { 0.14} &
  { 0.33} &
  { 0.32} &
  { 5.06} &
  { 4.85} \\
{ CPD\cite{wu2024commonsense}} &
  { CVPR2024} &
  { 37.40} &
  { 32.13} &
  { 16.31} &
  { 13.22} &
  { 5.06} &
  { 4.87} &
  { 19.59} &
  { 16.74} \\
{ OWL (Ours)} &
  { -} &
  { \textbf{48.08}} &
  { \textbf{41.51}} &
  { \textbf{29.66}} &
  { \textbf{24.68}} &
  { \textbf{32.27}} &
  { \textbf{31.14}} &
  { \textbf{36.91}} &
  { \textbf{32.44}} \\ \midrule
{ \textit{Improvement}} &
  { \textit{}} &
  { \textit{+10.68}} &
  { \textit{+9.38}} &
  { \textit{+13.35}} &
  { \textit{+11.46}} &
  { \textit{+27.21}} &
  { \textit{+26.27}} &
  { \textit{+17.32}} &
  { \textit{+15.70}} \\ \bottomrule
\end{tabularx}
\caption{Unsupervised 3D object detection results on WOD validation set. AP@0.7 means the average precision at $IoU_{0.7}$. The best performance is highlighted in bold.}
\label{tab1}
\end{table*}
\begin{table*}[htbp]
\centering
\setlength{\tabcolsep}{1.5mm}
\begin{tabularx}{\textwidth}{c|cccc|cccc|cccc}
\toprule
\multirow{3}{*}{Method} &
  \multicolumn{4}{c|}{Vehicle AP@0.7} &
  \multicolumn{4}{c|}{Pedestrian AP@0.5} &
  \multicolumn{4}{c}{Cyclist AP@0.5} \\
 &
  \multicolumn{2}{c}{L1} &
  \multicolumn{2}{c|}{L2} &
  \multicolumn{2}{c}{L1} &
  \multicolumn{2}{c|}{L2} &
  \multicolumn{2}{c}{L1} &
  \multicolumn{2}{c}{L2} \\
 &
  AP &
  APH &
  AP &
  APH &
  AP &
  APH &
  AP &
  APH &
  AP &
  APH &
  AP &
  APH \\ \midrule
MODEST &
  7.58 &
  7.22 &
  6.57 &
  6.26 &
  0.02 &
  0.01 &
  0.02 &
  0.01 &
  0.00 &
  0.00 &
  0.00 &
  0.00 \\
OYSTER &
  21.66 &
  20.81 &
  18.79 &
  18.05 &
  0.64 &
  0.32 &
  0.57 &
  0.28 &
  0.01 &
  0.00 &
  0.01 &
  0.00 \\
CPD &
  37.26 &
  35.00 &
  32.49 &
  30.51 &
  18.65 &
  8.34 &
  16.57 &
  7.40 &
  5.71 &
  3.69 &
  5.55 &
  3.55 \\
OWL (Ours) &
  \textbf{52.53} &
  \textbf{48.22} &
  \textbf{45.97} &
  \textbf{42.19} &
  \textbf{28.69} &
  \textbf{12.54} &
  \textbf{25.53} &
  \textbf{11.13} &
  \textbf{41.52} &
  \textbf{23.42} &
  \textbf{39.92} &
  \textbf{22.52} \\ \midrule
\textit{Improvement} &
  \textit{+15.27} &
  \textit{+13.22} &
  \textit{+13.48} &
  \textit{+11.68} &
  \textit{+10.04} &
  \textit{+4.20} &
  \textit{+8.96} &
  \textit{+3.73} &
  \textit{+35.81} &
  \textit{+19.73} &
  \textit{+34.37} &
  \textit{+18.97} \\ \midrule
\end{tabularx}
\caption{Unsupervised 3D object detection results on WOD test set. }
\label{tab2}
\end{table*}

\paragraph{Instance Cue  Mining.}
After label initialization, we obtain the initial pseudo-label $\mathcal{B} = \{b_j\}_j$, where $b_j = [x, y, z, l, w, h, \alpha, c]_j$ denotes position, length, width, height, orientation and class name respectively of the $j\ th$ box as the prior for the size cue. We apply category-agnostic tracking $\mathcal{T}$ to potential targets based on the detection boxes and adjust the inappropriate sizes of tracked objects and missed targets according to temporal consistency. We also estimate the motion of instances $v_{j}$ based on this, and this motion cue helps to distinguish between foreground and background targets. Subsequently, we calculate a variety of instance attributes $\mathcal{O}$ for the targets, including the number of points in the point cloud and the average intensity information. Finally, we calculate the prior reference confidence by distribution score $s_{dis}$ and consistency score $s_{cons}$ for each label:
\begin{align}
    s_{dis} &= [1-\mathcal{N}(\|(x_j, y_j, z_j)\|)] + \frac{N_j}{r\times r} , \\
    s_{cons} &= min(0.05,\sum_{k}\mathcal{S}_klog(\frac{\mathcal{S}_k}{s_k}))/0.05,
\end{align}
where $\mathcal{N}(\|(x_j, y_j, z_j)\|)$ denotes the distance normalization of $b_j$, $N_j$ is the number of grids occupied by points in the box, and $r$ is the resolution. $S_k$ is the size set for each category by LLMs, and $s_j=\{l_j,w_j,h_j\}$ denotes the length, width, and height of the detected bounding box, respectively. Unlike other methods, we use only this score as a cue for reference.

\paragraph{Large Model Priors Reasoning Refiner.}
After extracting a substantial amount of instance cues, we need to gauge the plausibility of each bounding box based on these cues. Large-scale reasoning models present a promising approach for this task. We feed the label information into a cue reasoner $\mathcal{G}$. For every bounding box $\mathcal{B} = \{b_j\}$, we prompt the discriminator to assess the validity of the pseudo-label:
\begin{equation}
    m_k,s_{rea},\Delta{L},cls_{new} = \mathcal{G}(\mathcal{B},\mathcal{P},\mathcal{V},\mathcal{O}) ,
\end{equation}
where $m_k$, $s_{\text{rea}}$, and $\Delta L$ denote the box mask, the reasoning score, and the correction vector for width, height, and length, respectively. And $\mathcal{P}$, $\mathcal{V}$, $\mathcal{O}$ denotes the prompt, motion cue and instance cue.

After cue reasoning, we perform filtering and refinement of the pseudo-labels. Bounding boxes $b_k$ that pass the filtering stage have their dimensions and class labels updated according to the network predictions. Predictions that are filtered out are nevertheless retained only if their consistency score exceeds a threshold $\eta$, yet their confidence weights are down-scaled, thereby reducing their contribution during training. $L_{com}$ denotes the average size of the corresponding category provided by the large model.
\begin{equation}
b_k' \;=\;
\begin{cases}
 b_k + \Delta L, & m_k = 1, \\
L_{com}, & m_k = 0 \;\wedge\; s_{cons} > \eta, \\
\varnothing, & {otherwise}.
\end{cases}
\end{equation}

\subsection{Weight-adapted Self-training}

After obtaining the filtered pseudo-labels, we train the detection network \( F \). Following the OGW module, we warm up the 3D backbone of the detector to guide the network toward preliminary learning of noise-free global features. Subsequently, following standard 3D detection practices, we use the pseudo-labels as supervisory signals to regress the object bounding boxes and predict their classes. The regression and classification tasks are optimized using the smooth absolute error loss and the focal loss\cite{lin2017focal}, respectively:

\begin{equation}
    \mathcal{L}_{\text{total}}=\frac{1}{Nf}\sum_{i}\omega_{i}[\alpha\mathcal{L}_{\text{reg}}+\beta\mathcal{L}_{\text{cls}}],
\end{equation}
where \(\omega_{i}\) denotes the weight assigned to each object’s pseudo-label, reflecting its quality:

\begin{equation}
   \omega_{i}=\lambda_{1}s_{\text{cons}}+\lambda_{2}s_{\text{rea}},
\end{equation}
$N_f$, $\mathcal{L}_{reg}$, and $\mathcal{L}_{cls}$ denote the number of samples, regression loss, and classification loss. $\lambda_{1}=\lambda_{2}=1$, $\alpha=2\beta=1$.

After several training epochs, we perform test-time augmentation (TTA) inference on the training set using the current detector. The resulting predictions undergo a similar filtering and refinement process to generate the initial labels \(\mathcal{B}_{\text{ST1}}\) for the next round. 

\begin{table}[ht]
\centering
\setlength{\tabcolsep}{1.2mm}
\begin{tabularx}{0.47\textwidth}{@{}c|ccc@{}}
\toprule
\multirow{2}{*}{Method} & \multicolumn{3}{c}{BEV AP}                       \\
                        & Vehicle@0.7    & Pedestrian@0.5 & Cyclist@0.5    \\ \midrule
MODEST                  & 13.87          & 0.10           & 0.30           \\
OYSTER                  & 23.67          & 0.20           & 0.10           \\
CPD                     & 51.56          & 17.84          & 4.95           \\
OWL (Ours)           & \textbf{57.76} & \textbf{30.85} & \textbf{33.72} \\ \bottomrule
\end{tabularx}
\caption{BEV results on WOD validation set.}
\label{tab3}
\end{table}

\begin{table*}[!htbp]
\centering
\begin{tabularx}{0.98\textwidth}{c|c|c|ccc|ccc}
\midrule
 &
   &
   &
  \multicolumn{3}{c|}{3D AP} &
  \multicolumn{3}{c}{3D AP} \\
\multirow{-2}{*}{Method} &
  \multirow{-2}{*}{Setting} &
  \multirow{-2}{*}{Label rate} &
  { Veh.@0.7} &
  { Ped.@0.5} &
  { Cyc.@0.5} &
  { Veh.@0.4} &
  { Ped.@0.4} &
  { Cyc.@0.4} \\ \midrule
{ \textit{Centerpoint}} &
  { Fully.} &
  { 100\%} &
  { 67.25} &
  { 68.18} &
  { 67.40} &
  { 89.07} &
  { 80.50} &
  { 76.49} \\ \midrule
{ \textit{CoIn}} &
  { } &
  { } &
  { 45.04} &
  { 26.02} &
  { 62.85} &
  { -} &
  { -} &
  { -} \\
{ \textit{SP3D}} &
  \multirow{-2}{*}{{ Weakly.}} &
  \multirow{-2}{*}{{ 2\%}} &
  { 43.54} &
  { 13.27} &
  { 60.65} &
  { -} &
  { -} &
  { -} \\ \midrule
{ MODEST} &
  { } &
  { } &
  { 6.13} &
  { 0.14} &
  { 1.12} &
  { 21.50} &
  { 0.50} &
  { 1.40} \\
{ OYSTER} &
  { } &
  { } &
  { 14.38} &
  { 0.16} &
  { 0.33} &
  { 32.80} &
  { 0.20} &
  { 0.00} \\
{ LISO} &
  { } &
  { } &
  { -} &
  { -} &
  { -} &
  { 54.30} &
  { 3.70} &
  { 1.60} \\
{ CPD} &
  { } &
  { } &
  { 34.77} &
  { 14.77} &
  { 4.970} &
  { 58.90} &
  { 19.20} &
  { 5.30} \\
OWL (Ours) &
  \multirow{-5}{*}{{ Unsup.}} &
  \multirow{-5}{*}{{ 0\%}} &
  { \textbf{44.80}} &
  { \textbf{27.17}} &
  { \textbf{31.71}} &
  { \textbf{64.90}} &
  { \textbf{34.64}} &
  { \textbf{34.24}} \\ \midrule
\end{tabularx}
\caption{Comparison with weakly/fully supervised and unsupervised methods on WOD validation set. 'Fully.', 'Weakly.', and 'Unsup.' mean fully supervised, weakly supervised, and unsupervised settings, respectively.}
\label{tab4}
\end{table*}
\begin{table*}[!ht]
\centering
\setlength{\tabcolsep}{3mm}
\begin{tabularx}{\textwidth}{@{}c|c|c|c|ccc|ccc@{}}
\toprule
{ } &
   &
  { } &
  { } &
  \multicolumn{3}{c|}{{ 3D AP@IoU0.5}} &
  \multicolumn{3}{c}{{ Car 3D AP@IoU0.7}} \\ 
\multirow{-2}{*}{{ Method}} &
  \multirow{-2}{*}{Modality} &
  \multirow{-2}{*}{{ Setting}} &
  \multirow{-2}{*}{{ Label ratio}} &
  { Car} &
  { Ped.} &
  { Cyc.} &
  { Easy} &
  { Mod.} &
  { Hard} \\ \midrule
{ \textit{Centerpoint}} &
  L &
  { } &
  { } &
  { 94.77} &
  { 52.11} &
  { 69.32} &
  { 89.07} &
  { 80.50} &
  { 76.49} \\
{ \textit{Voxel-RCNN}} &
  L &
  \multirow{-2}{*}{{ Fully Sup.}} &
  \multirow{-2}{*}{{ 100\%}} &
  { 96.88} &
  { 63.73} &
  { 76.47} &
  { 92.38} &
  { 85.29} &
  { 82.86} \\ \midrule
{ \textit{CoIn}} &
  L &
  { } &
  { } &
  { 83.89} &
  { 31.69} &
  { 49.10} &
  { 89.17} &
  { 75.32} &
  { 62.98} \\
{ \textit{SP3D}} &
  L+I &
  \multirow{-2}{*}{{ Weakly Sup.}} &
  \multirow{-2}{*}{{ 2\%}} &
  { 80.37} &
  { 59.10} &
  { 77.91} &
  { 91.12} &
  { 75.94} &
  { 66.46} \\ \midrule
{ MODEST} &
  L &
  { } &
  { } &
  { 37.19} &
  { 1.96} &
  { 0.10} &
  { 12.65} &
  { 11.14} &
  { 10.60} \\
{ OYSTER} &
  L &
  { } &
  { } &
  { 54.58} &
  { 3.33} &
  { 1.76} &
  { 23.22} &
  { 20.31} &
  { 19.97} \\
{ LISO} &
  L &
  { } &
  { } &
  { 41.10} &
  { 9.70} &
  { 5.30} &
  { -} &
  { -} &
  { -} \\
{ CPD} &
  L &
  { } &
  { } &
  { 83.89} &
  { 15.48} &
  { 8.30} &
  { 72.98} &
  { 55.07} &
  { 53.94} \\
OWL (Ours) &
  L &
  \multirow{-5}{*}{{ Unsup.}} &
  \multirow{-5}{*}{{ 0\%}} &
  { \textbf{86.10}} &
  { \textbf{30.23}} &
  { \textbf{50.98}} &
  { \textbf{79.80}} &
  { \textbf{62.67}} &
  { \textbf{58.11}} \\ \bottomrule
\end{tabularx}
\caption{Comparison with weakly/fully supervised and unsupervised methods on KITTI validation set. 'L' and 'I' denote lidar and image modality.}
\label{tab5}
\end{table*}
\section{Experiments}

\subsection{Datasets and Metrics}
\paragraph{KITTI Dataset.}

The KITTI \cite{geiger2012we} dataset covers diverse scenarios such as urban, rural, and highway environments, and is one of the earliest and most widely used datasets in autonomous driving. It is divided into 3,712 frames for training and 3,769 for validation. 

\paragraph{Waymo Open Dataset (WOD).}
The WOD \cite{sun2020scalability} consists of 5 RGB cameras and 5 LiDAR sensors, capturing 1,150 scenes, each lasting 20 seconds, and containing over 200k frames of diverse and complex scene data. The WOD is divided into 798 training, 202 validation, and 150 testing sequences, respectively. No annotations were used during the training process.


\subsection{Comparison with SoTA Methods}
\paragraph{Comparison with Unsupervised Methods.}
\begin{figure}[!htbp]
\centering
\includegraphics[scale=0.24]{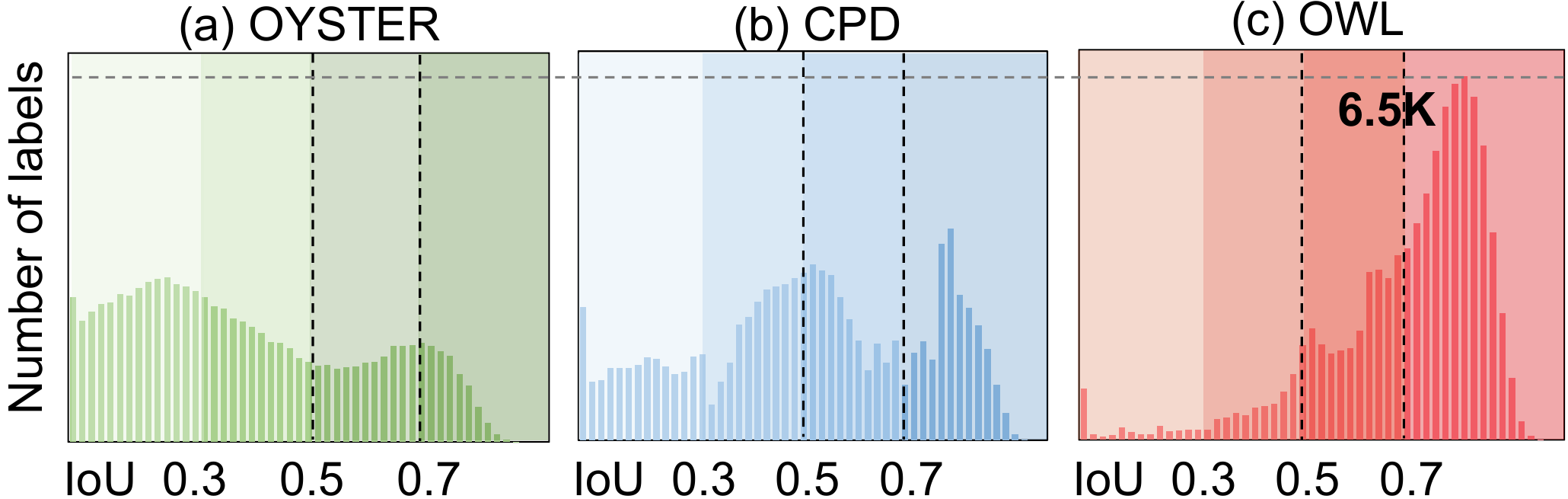} 
\caption{The IoU distribution between pseudo-labels and ground truth is presented.}
\label{fig4}
\end{figure}

The comparison results between OWL and unsupervised methods on the WOD validation set and test set are presented in Table \ref{tab1} and \ref{tab2}, respectively. The results demonstrate that our approach significantly outperforms prior unsupervised SoTA methods. For instance, on WOD validation set, OWL achieves relative improvements of 17.32\% and 15.70\% in L1 and L2 mAP over CPD. We also report the results of BEV AP on WOD in Table \ref{tab3}. This improvement derives from the design of our ICR, OGW, and WAS modules, which generate more accurate pseudo-labels and a more robust network initialization. 

\begin{table}[h]
\setlength{\tabcolsep}{2.4mm}
\begin{tabularx}{0.46\textwidth}{l|cc|cc}
\midrule
\multicolumn{1}{c|}{\multirow{2}{*}{Method}} & \multicolumn{2}{c|}{3D Recall}                                            & \multicolumn{2}{c}{3D Precision}                                         \\
\multicolumn{1}{c|}{}                        & \multicolumn{1}{c}{$IoU_{0.5}$} & \multicolumn{1}{c|}{$IoU_{0.7}$} & \multicolumn{1}{c}{$IoU_{0.5}$} & \multicolumn{1}{c}{$IoU_{0.7}$} \\ \midrule
MODEST                                       & 12.04                  & 4.89                    & 22.81                  & 10.05                  \\
OYSTER                                       & 21.01                  & 11.12                   & 21.09                  & 9.45                   \\
CPD                                          & 39.33                  & 20.54                   & 28.22                  & 14.74                  \\
OWL (Ours)                                & \textbf{56.77}                  & \textbf{33.19}                   & \textbf{37.35}                  & \textbf{20.16}                  \\ \midrule
\end{tabularx}
\caption{Pseudo-label comparison results on WOD validation set.}
\label{tab6}
\end{table}

\paragraph{Comparison with Fully/weakly Methods.}
As KITTI lacks continuous-frame sequences, we follow CPD’s protocol: we train on the WOD dataset and evaluate on the KITTI validation set, comparing with a variety of unsupervised and weakly-supervised methods. As shown in Table \ref{tab4} and \ref{tab5}, OWL surpasses prior unsupervised methods on both KITTI and WOD over 15\% mAP, outperforms several weakly-supervised and multimodal approaches, and approaches fully-supervised performance on KITTI, demonstrating the advancement of our method.

\begin{figure*}[!htbp]
    \centering
    \includegraphics[scale=0.35]{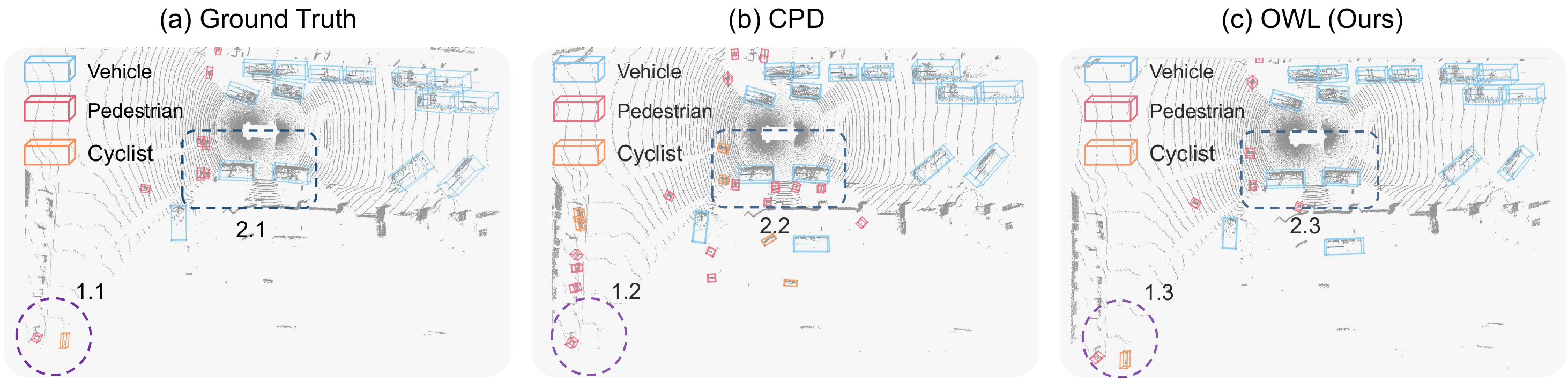} 
    \caption{Visualization comparison of different methods.}
    \label{fig5}
\end{figure*}
\paragraph{Pseudo-label Comparison.}
To evaluate the quality of pseudo-labels, we analyzed OWL’s Recall and Precision performance on the WOD validation set. As shown in Table \ref{tab6}, OWL boosts Recall by 12.65\% and Precision by 5.42\% at $IoU_{0.7}$. Figure \ref{fig4} displays the IoU distribution between our pseudo-labels and the ground truth, demonstrating that our pseudo-labels outperform those of prior methods. This improvement stems from the effective screening and refinement performed by our ICR module.

\begin{table}[h]
\centering
\setlength{\tabcolsep}{2.1mm}
\begin{tabularx}{0.46\textwidth}{cccc|c|c}
\midrule
\multicolumn{4}{c|}{Components}                   & \multirow{2}{*}{3D AP L1} & \multirow{2}{*}{3D AP L2} \\
PLI        & ICR      & OGW       & WAS        &                           &                           \\ \midrule
\textbf{\checkmark} & \textbf{}  & \textbf{}  & \textbf{}  & 19.59                     & 16.74                     \\
\textbf{\checkmark} & \textbf{\checkmark} & \textbf{}  & \textbf{}  & 30.25                    & 26.38                     \\
\textbf{\checkmark} & \textbf{\checkmark} & \textbf{\checkmark} & \textbf{}  & 34.16                    & 30.81                    \\
\textbf{\checkmark} & \textbf{\checkmark} & \textbf{\checkmark} & \textbf{\checkmark} & \textbf{36.67}            & \textbf{32.44}            \\ \midrule
\end{tabularx}
\caption{Ablation study of our method.}
\label{tab7}
\end{table}
\begin{table}[htbp]
\centering
\setlength{\tabcolsep}{2.2mm}
\begin{tabularx}{0.465\textwidth}{c|cccc}
\toprule
\multirow{2}{*}{Warm-up methods} & \multicolumn{2}{c}{L1}          & \multicolumn{2}{c}{L2}          \\
                                 & AP             & APH            & AP             & APH            \\ \midrule
Train directly                   & 30.25          & 24.01          & 26.38          & 21.16          \\
Contrastive learning             & 29.10          & 23.98          & 25.54          & 20.97          \\
Point completion                 & 31.06          & 24.65          & 27.08          & 21.53          \\
OGW                              & \textbf{34.16} & \textbf{26.63} & \textbf{30.81} & \textbf{23.40} \\ \midrule
\end{tabularx}
\caption{Comparison with different warm-up methods on WOD validation set.}
\label{tab8}
\end{table}

\subsection{Ablation Study}

\paragraph{Components analysis of OWL.}
We evaluated each component and assessed its impact using the WOD validation set, with the results shown in Table \ref{tab7}. Compared with previous pseudo-label initialization (PLI), our ICR module yields a 10.66\% AP improvement, attributed to leveraging large models and instance-level cues to filter and refine a substantial number of erroneous pseudo-labels. OGW further improves detection performance by 3.91\% by guiding stable initialization, mitigating the misleading impact of incorrect labels, and enhancing spatial awareness.

\paragraph{Influence of different warm-up methods.  }
As shown in Table \ref{tab8}, we examine the impact of different warm-up strategies on detector performance. Compared to direct training, contrastive learning underperforms because it typically relies on labels to distinguish between positive and negative samples, whereas point completion yields only a limited gain. The experiment confirms the effectiveness of OGW.

\paragraph{Performance along different distances. } 
\begin{table}[htbp]
\centering
\setlength{\tabcolsep}{4.3mm}
\begin{tabularx}{0.465\textwidth}{@{}c|ccc@{}}
\toprule
\multirow{2}{*}{Method} & \multicolumn{3}{c}{3D mAP}               \\
                        & {[}0,30) m & {[}30,50) m & {[}50,inf{)}  \\ \midrule
OYSTER                  & 15.30      & 3.26        & 0.28          \\
CPD                     & 31.49      & 17.84       & 5.14          \\
OWL (Ours)                    & \textbf{57.74}      & \textbf{37.16}       & \textbf{13.56}         \\ \bottomrule
\end{tabularx}
\caption{Range detection results on WOD validation set.}
\label{tab9}
\end{table}
Table \ref{tab9} presents the performance across varying distances, showing that OWL achieves superior object detection at longer ranges. 

\subsection{Visualization Analysis}

In Figure \ref{fig5}, OWL's improvements are visually demonstrated. Through label filtering and screening, OWL significantly reduces pseudo-label annotation errors, as shown by the marked decrease in False-Negative and False-Positive samples in Figure \ref{fig5} (2.2, 2.3). Additionally, detection performance for under-represented categories, like Cyclist in Figure \ref{fig5} (1.3), is significantly enhanced.

\section{Conclusion}

In this paper, we propose OWL, a new framework for enhancing unsupervised 3D object detection. We introduce the OGW module to guide network initialization, mitigating the impact of incorrect labels during early training stages. To address the prevalence of noisy pseudo-labels, we propose the ICR module to extract instance cues and leverage the knowledge and reasoning capabilities of large models to refine the pseudo-labels. Finally, we design the WAS strategy to learn the loss adaptively. These designs mutually reinforce pseudo-label quality to improve the detection performance.
\paragraph{Limitations.} OWL uses LLM to refine labels, but this process depends on the model's performance. Most models lack point cloud data and may produce hallucinations. Future work could explore using VLMs for better reasoning.


\section{Acknowledgements}This work was supported in part
 by the National Natural Science Foundation of China
 (No.42571514).



\bigskip
\noindent 

\bibliography{main}

@inproceedings{deng2021voxel,
  title={Voxel r-cnn: Towards high performance voxel-based 3d object detection},
  author={Deng, Jiajun and Shi, Shaoshuai and Li, Peiwei and Zhou, Wengang and Zhang, Yanyong and Li, Houqiang},
  booktitle={Proceedings of the AAAI conference on artificial intelligence},
  volume={35},
  number={2},
  pages={1201--1209},
  year={2021}
}

@inproceedings{yin2021center,
  title={Center-based 3d object detection and tracking},
  author={Yin, Tianwei and Zhou, Xingyi and Krahenbuhl, Philipp},
  booktitle={Proceedings of the IEEE/CVF conference on computer vision and pattern recognition},
  pages={11784--11793},
  year={2021}
}

@article{wu2022casa,
  title={CasA: A cascade attention network for 3-D object detection from LiDAR point clouds},
  author={Wu, Hai and Deng, Jinhao and Wen, Chenglu and Li, Xin and Wang, Cheng and Li, Jonathan},
  journal={IEEE Transactions on Geoscience and Remote Sensing},
  volume={60},
  pages={1--11},
  year={2022},
  publisher={IEEE}
}

@inproceedings{wu2023virtual,
  title={Virtual sparse convolution for multimodal 3d object detection},
  author={Wu, Hai and Wen, Chenglu and Shi, Shaoshuai and Li, Xin and Wang, Cheng},
  booktitle={Proceedings of the IEEE/CVF conference on computer vision and pattern recognition},
  pages={21653--21662},
  year={2023}
}

@inproceedings{wu2023transformation,
  title={Transformation-equivariant 3d object detection for autonomous driving},
  author={Wu, Hai and Wen, Chenglu and Li, Wei and Li, Xin and Yang, Ruigang and Wang, Cheng},
  booktitle={Proceedings of the AAAI Conference on Artificial Intelligence},
  volume={37},
  number={3},
  pages={2795--2802},
  year={2023}
}

@inproceedings{zhao2025sp3d,
  title={SP3D: Boosting Sparsely-Supervised 3D Object Detection via Accurate Cross-Modal Semantic Prompts},
  author={Zhao, Shijia and Xia, Qiming and Guo, Xusheng and Zou, Pufan and Zheng, Maoji and Wu, Hai and Wen, Chenglu and Wang, Cheng},
  booktitle={Proceedings of the Computer Vision and Pattern Recognition Conference},
  pages={29374--29384},
  year={2025}
}

@inproceedings{wu2024commonsense,
  title={Commonsense prototype for outdoor unsupervised 3d object detection},
  author={Wu, Hai and Zhao, Shijia and Huang, Xun and Wen, Chenglu and Li, Xin and Wang, Cheng},
  booktitle={Proceedings of the IEEE/CVF Conference on Computer Vision and Pattern Recognition},
  pages={14968--14977},
  year={2024}
}

@inproceedings{xia2023coin,
  title={Coin: Contrastive instance feature mining for outdoor 3d object detection with very limited annotations},
  author={Xia, Qiming and Deng, Jinhao and Wen, Chenglu and Wu, Hai and Shi, Shaoshuai and Li, Xin and Wang, Cheng},
  booktitle={Proceedings of the IEEE/CVF International Conference on Computer Vision},
  pages={6254--6263},
  year={2023}
}

@inproceedings{baur2024liso,
  title={Liso: Lidar-only self-supervised 3d object detection},
  author={Baur, Stefan Andreas and Moosmann, Frank and Geiger, Andreas},
  booktitle={European Conference on Computer Vision},
  pages={253--270},
  year={2024},
  organization={Springer}
}

@inproceedings{ester1996density,
  title={A density-based algorithm for discovering clusters in large spatial databases with noise},
  author={Ester, Martin and Kriegel, Hans-Peter and Sander, J{\"o}rg and Xu, Xiaowei and others},
  booktitle={kdd},
  volume={96},
  number={34},
  pages={226--231},
  year={1996}
}

@inproceedings{you2022learning,
  title={Learning to detect mobile objects from lidar scans without labels},
  author={You, Yurong and Luo, Katie and Phoo, Cheng Perng and Chao, Wei-Lun and Sun, Wen and Hariharan, Bharath and Campbell, Mark and Weinberger, Kilian Q},
  booktitle={Proceedings of the IEEE/CVF Conference on Computer Vision and Pattern Recognition},
  pages={1130--1140},
  year={2022}
}

@inproceedings{zhang2023towards,
  title={Towards unsupervised object detection from lidar point clouds},
  author={Zhang, Lunjun and Yang, Anqi Joyce and Xiong, Yuwen and Casas, Sergio and Yang, Bin and Ren, Mengye and Urtasun, Raquel},
  booktitle={Proceedings of the IEEE/CVF Conference on Computer Vision and Pattern Recognition},
  pages={9317--9328},
  year={2023}
}

@article{min2023occupancy,
  title={Occupancy-mae: Self-supervised pre-training large-scale lidar point clouds with masked occupancy autoencoders},
  author={Min, Chen and Xiao, Liang and Zhao, Dawei and Nie, Yiming and Dai, Bin},
  journal={IEEE Transactions on Intelligent Vehicles},
  volume={9},
  number={7},
  pages={5150--5162},
  year={2023},
  publisher={IEEE}
}

@inproceedings{zheng2025seg2box,
  title={Seg2Box: 3D Object Detection by Point-Wise Semantics Supervision},
  author={Zheng, Maoji and Xu, Ziyu and Xia, Qiming and Wu, Hai and Wen, Chenglu and Wang, Cheng},
  booktitle={Proceedings of the AAAI Conference on Artificial Intelligence},
  volume={39},
  number={10},
  pages={10591--10598},
  year={2025}
}

@inproceedings{geiger2012we,
  title={Are we ready for autonomous driving? the kitti vision benchmark suite},
  author={Geiger, Andreas and Lenz, Philip and Urtasun, Raquel},
  booktitle={2012 IEEE conference on computer vision and pattern recognition},
  pages={3354--3361},
  year={2012},
  organization={IEEE}
}

@inproceedings{sun2020scalability,
  title={Scalability in perception for autonomous driving: Waymo open dataset},
  author={Sun, Pei and Kretzschmar, Henrik and Dotiwalla, Xerxes and Chouard, Aurelien and Patnaik, Vijaysai and Tsui, Paul and Guo, James and Zhou, Yin and Chai, Yuning and Caine, Benjamin and others},
  booktitle={Proceedings of the IEEE/CVF conference on computer vision and pattern recognition},
  pages={2446--2454},
  year={2020}
}

@inproceedings{cmd,
  title={Cmd: A cross mechanism domain adaptation dataset for 3d object detection},
  author={Deng, Jinhao and Ye, Wei and Wu, Hai and Huang, Xun and Xia, Qiming and Li, Xin and Fang, Jin and Li, Wei and Wen, Chenglu and Wang, Cheng},
  booktitle={European Conference on Computer Vision},
  pages={219--236},
  year={2024},
  organization={Springer}
}

@inproceedings{l4dr,
  title={L4dr: Lidar-4dradar fusion for weather-robust 3d object detection},
  author={Huang, Xun and Xu, Ziyu and Wu, Hai and Wang, Jinlong and Xia, Qiming and Xia, Yan and Li, Jonathan and Gao, Kyle and Wen, Chenglu and Wang, Cheng},
  booktitle={Proceedings of the AAAI Conference on Artificial Intelligence},
  volume={39},
  number={4},
  pages={3806--3814},
  year={2025}
}

@article{guo2022hssda,
  title={HSSDA: Hierarchical relation aided Semi-Supervised Domain Adaptation},
  author={Guo, Xiechao and Liu, Ruiping and Song, Dandan},
  journal={AI Open},
  volume={3},
  pages={156--161},
  year={2022},
  publisher={Elsevier}
}

@inproceedings{dsvt,
  title={Dsvt: Dynamic sparse voxel transformer with rotated sets},
  author={Wang, Haiyang and Shi, Chen and Shi, Shaoshuai and Lei, Meng and Wang, Sen and He, Di and Schiele, Bernt and Wang, Liwei},
  booktitle={Proceedings of the IEEE/CVF Conference on Computer Vision and Pattern Recognition},
  pages={13520--13529},
  year={2023}
}

@inproceedings{ss3d,
  title={Ss3d: Sparsely-supervised 3d object detection from point cloud},
  author={Liu, Chuandong and Gao, Chenqiang and Liu, Fangcen and Liu, Jiang and Meng, Deyu and Gao, Xinbo},
  booktitle={Proceedings of the IEEE/CVF conference on computer vision and pattern recognition},
  pages={8428--8437},
  year={2022}
}

@article{hass,
  title={Hardness-aware scene synthesis for semi-supervised 3D object detection},
  author={Zeng, Shuai and Zheng, Wenzhao and Lu, Jiwen and Yan, Haibin},
  journal={IEEE Transactions on Multimedia},
  volume={26},
  pages={9644--9656},
  year={2024},
  publisher={IEEE}
}

@inproceedings{hinted,
  title={Hinted: Hard instance enhanced detector with mixed-density feature fusion for sparsely-supervised 3d object detection},
  author={Xia, Qiming and Ye, Wei and Wu, Hai and Zhao, Shijia and Xing, Leyuan and Huang, Xun and Deng, Jinhao and Li, Xin and Wen, Chenglu and Wang, Cheng},
  booktitle={Proceedings of the IEEE/CVF Conference on Computer Vision and Pattern Recognition},
  pages={15321--15330},
  year={2024}
}

@inproceedings{shi2019pointrcnn,
  title={Pointrcnn: 3d object proposal generation and detection from point cloud},
  author={Shi, Shaoshuai and Wang, Xiaogang and Li, Hongsheng},
  booktitle={Proceedings of the IEEE/CVF conference on computer vision and pattern recognition},
  pages={770--779},
  year={2019}
}

@article{lentsch2024union,
  title={Union: Unsupervised 3d object detection using object appearance-based pseudo-classes},
  author={Lentsch, Ted and Caesar, Holger and Gavrila, Dariu},
  journal={Advances in Neural Information Processing Systems},
  volume={37},
  pages={22028--22046},
  year={2024}
}

@article{gidaris2018unsupervised,
  title={Unsupervised representation learning by predicting image rotations},
  author={Gidaris, Spyros and Singh, Praveer and Komodakis, Nikos},
  journal={arXiv preprint arXiv:1803.07728},
  year={2018}
}

@inproceedings{xie2020pointcontrast,
  title={Pointcontrast: Unsupervised pre-training for 3d point cloud understanding},
  author={Xie, Saining and Gu, Jiatao and Guo, Demi and Qi, Charles R and Guibas, Leonidas and Litany, Or},
  booktitle={European conference on computer vision},
  pages={574--591},
  year={2020},
  organization={Springer}
}

@inproceedings{he2022masked,
  title={Masked autoencoders are scalable vision learners},
  author={He, Kaiming and Chen, Xinlei and Xie, Saining and Li, Yanghao and Doll{\'a}r, Piotr and Girshick, Ross},
  booktitle={Proceedings of the IEEE/CVF conference on computer vision and pattern recognition},
  pages={16000--16009},
  year={2022}
}

@inproceedings{hess2023masked,
  title={Masked autoencoder for self-supervised pre-training on lidar point clouds},
  author={Hess, Georg and Jaxing, Johan and Svensson, Elias and Hagerman, David and Petersson, Christoffer and Svensson, Lennart},
  booktitle={Proceedings of the IEEE/CVF winter conference on applications of computer vision},
  pages={350--359},
  year={2023}
}

@inproceedings{xu2023hypermodest,
  title={HyperMODEST: Self-Supervised 3D Object Detection with Confidence Score Filtering},
  author={Xu, Jenny and Waslander, Steven L},
  booktitle={2023 20th Conference on Robots and Vision (CRV)},
  pages={217--224},
  year={2023},
  organization={IEEE}
}

@inproceedings{qi2017pointnet,
  title={Pointnet: Deep learning on point sets for 3d classification and segmentation},
  author={Qi, Charles R and Su, Hao and Mo, Kaichun and Guibas, Leonidas J},
  booktitle={Proceedings of the IEEE conference on computer vision and pattern recognition},
  pages={652--660},
  year={2017}
}

@inproceedings{lin2017focal,
  title={Focal loss for dense object detection},
  author={Lin, Tsung-Yi and Goyal, Priya and Girshick, Ross and He, Kaiming and Doll{\'a}r, Piotr},
  booktitle={Proceedings of the IEEE international conference on computer vision},
  pages={2980--2988},
  year={2017}
}

@inproceedings{wu2025motal,
  title={Motal: Unsupervised 3D Object Detection by Modality and Task-specific Knowledge Transfer},
  author={Wu, Hai and Lin, Hongwei and Guo, Xusheng and Li, Xin and Wang, Mingming and Wang, Cheng and Wen, Chenglu},
  booktitle={Proceedings of the IEEE/CVF International Conference on Computer Vision},
  pages={6284--6293},
  year={2025}
}
\end{document}